\documentclass[letterpaper]{article} 
\usepackage{aaai25}

\usepackage{times}  
\usepackage{helvet}  
\usepackage{courier}  
\usepackage[hyphens]{url}  
\usepackage{graphicx} 
\usepackage{natbib}  
\usepackage{caption} 
\usepackage{algorithm}
\usepackage{algorithmic}

\usepackage{newfloat}
\usepackage{listings}
\DeclareCaptionStyle{ruled}{labelfont=normalfont,labelsep=colon,strut=off} 
\floatstyle{ruled}
\newfloat{listing}{tb}{lst}{}
\floatname{listing}{Listing}
\title{Psy-Copilot: Visual Chain of Thought for Counseling}
\author{
    Keqi Chen,
    Zekai Sun,
    Huijun Lian,
    Yingming Gao,
    Ya Li
}
\affiliations{

    School of Artificial Intelligence\\
    Beijing University of Posts and Telecommunications\\
}

\usepackage{bibentry}

\begin{document}

\maketitle

\begin{abstract}

Large language models (LLMs) are becoming increasingly popular in the field of psychological counseling. However, when human therapists work with LLMs in therapy sessions, it is hard to understand how the model gives the answers. To address this, we have constructed Psy-COT, a graph designed to visualize the thought processes of LLMs during therapy sessions.  The Psy-COT graph presents semi-structured counseling conversations alongside step-by-step annotations that capture the reasoning and insights of therapists. Moreover, we have developed Psy-Copilot, which is a conversational AI assistant designed to assist human psychological therapists in their consultations. It can offer traceable psycho-information based on retrieval, including response candidates, similar dialogue sessions, related strategies, and visual traces of results. We have also built an interactive platform for AI-assisted counseling. It has an interface that displays the relevant parts of the retrieval sub-graph. The Psy-Copilot is designed not to replace psychotherapists, but to foster collaboration between AI and human therapists, thereby promoting mental health development. Our code and demo are both open-sourced and available for use. Users can act as clients or therapists to interact with Psy-Copilot on this website \footnote{\url{https://ckqqqq.github.io/Demo/Psy-Copilot/}}.
\end{abstract}

%

\section{Introduction}

In recent years, empathetic conversational agents that can provide psychological support have become increasingly popular. However, most existing mental support dialogue agents are designed to interact directly with clients, lacking the ability to collaborate with psychotherapists in counseling~\cite{CAiRE,Emohha,zhang2024cpsycoun}. In psychological counseling, human therapists need a copilot to assist in consultations, rather than assigning patients directly to a black-box AI conversational agent. Psychological copilot should not only provide response candidates but, more importantly, offer traceable and reliable information for therapists.

To achieve this goal, we construct a Psy-COT graph and Psy-Copilot conversational agent. They constitute an AI assistant for psychological counseling tasks. The Psy-COT graph consists of 941 embedded dialogue sessions and a directed graph of events and strategies. Psy-COT is a multi-level, indexable graph, which is designed for visualizing and retrieving the chain of thought progress in counseling. 

\begin{figure}[th]
    \centering
     \includegraphics[width=1\linewidth]{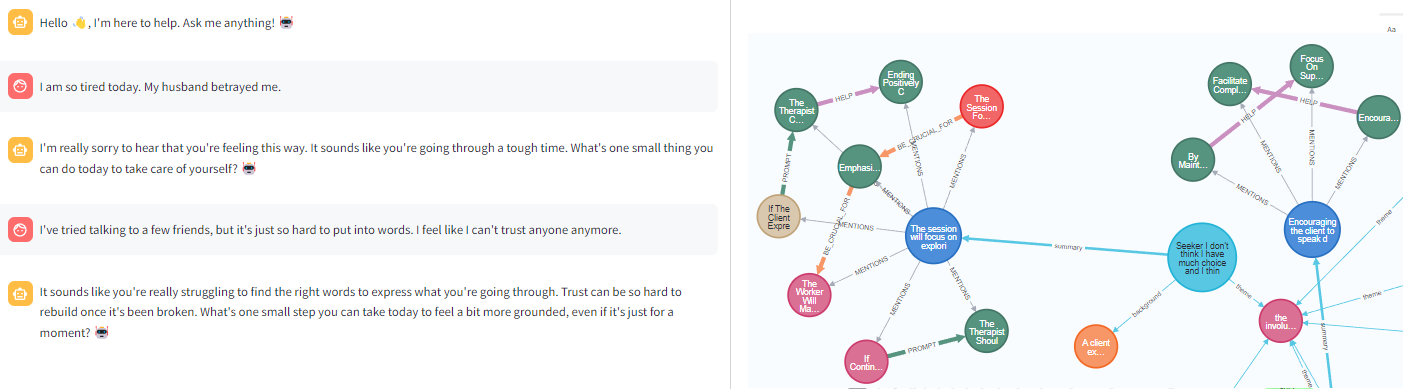}
    \caption{Screenshot of Psy-Copilot. The left side is the chat interface, and the right side is the visualized COT.}
    \label{fig:screenshot}
    \includegraphics[width=\linewidth]{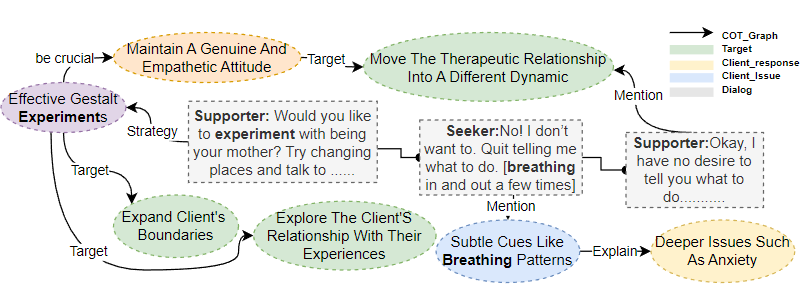}
    \caption{A small sub-graph of Psy-COT graph. The Psy-COT maps events and strategies to dialogue units, preserving causal and temporal relationships.}
    \label{fig:example}

\end{figure}

As shown in Figure~\ref{fig:screenshot}, Psy-Copilot is a conversation assistant that can aid therapists in counseling. It can leverage retrieval text to generate traceable counseling responses. With indexes in Psy-COT, Psy-Copilot can retrieve precise similar study cases and required strategies for reasoning. Psy-Copilot can enhance the performance of LLMs and provide better visualization of how the AI agent generates response candidates.

Psy-COT has similarities to knowledge graphs~\cite{Knowledgegraph}. As shown in Figure~\ref{fig:example}, unlike knowledge graphs, which focus on inclusion relationships between entities (eg., “Disappointed"), Psy-COT emphasizes the logical chain of causality in events and the temporal evolution of strategies in counseling (eg., Expand Client's Client's relationship). 

In our research, we focus on (1) Collecting counseling dialogue and interpretation data to build a graph that maps out event connections and strategic transitions; (2) Constructing an AI agent that can produce reliable, explainable, and traceable results; (3) Building an interactive platform to ensure that the AI agent can collaborate with human therapists, with outcomes that are visualized and easily comprehensible.




\section{Technical Details}
\begin{figure}
    \centering
    \includegraphics[width=\linewidth]{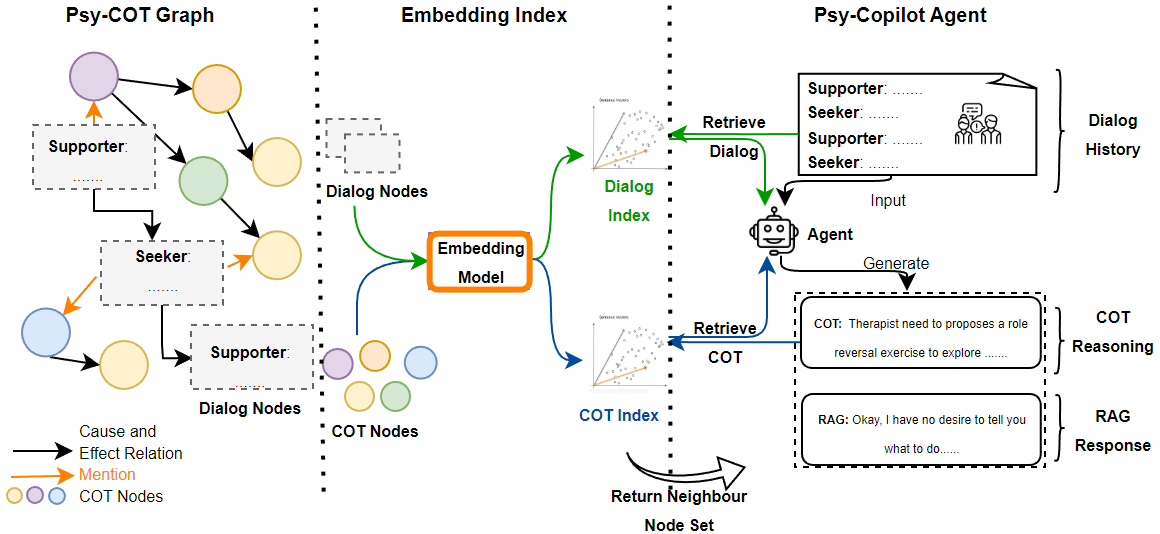}
    \caption{Overview of indexes in Psy-COT and retrieval progress in Psy-Copilot. Psy-COT has two vector indexes for dialog and COT reasoning content respectively. There is a two-stage retrieval argument generation in Psy-Copilot.  }
    \label{fig:overview}
\end{figure}
\subsection{Psy-COT Graph Construction}


\subsubsection{Data Source}

We collect 941 psycho-counseling sessions from psychological blogs, each containing dialog and corresponding explanatory text.

\subsubsection{Construction of COT Graph}
We design a multi-agent framework to automate the extraction of strategy descriptions from unstructured data. Based on psychological counseling research ~\cite{2015psychological-strategy}, we define the types of strategies and causal relationships. Using a Deepseek-powered Langchain Agent, we extract text fragments describing strategies and established causal relationships between nodes. Subsequently, we employ a sliding window technique to map these COT nodes to dialogue nodes, thereby establishing temporal relationships.

\subsubsection{Multi-Level Index for Retrieval}
In contrast to graphGAG~\cite{2024graphrag}, which employs embedded text chunks for document retrieval, we adopt a fine-grained approach that is designed for counseling dialog.
As shown in the left side of Figure \ref{fig:overview}, we construct two index structures for COT nodes and dialogue nodes. 

We have found the distinction between the content of the chain of thought and the dialogue. The COT content is derived from explanations in blogs, using specialized psychological terms and direct descriptions, such as "Effective Gestalt Experiments" or "Deeper Issues Such As Anxiety." However, the dialogue content is more euphemistic and, like, "Alright, I'm not here to tell you what to do." Given these differences, we have created separate indexes in Psy-COT for each content type, which makes vector retrieval more precise.

\subsection{Psy-Copilot Agent}

\subsubsection{Base LLMs}

We utilize open-source models Qwen-2.5-7B~\citeyearpar{bai2023qwentechnicalreport} and Deepseek-V2~\citeyearpar{deepseekai202} to power our retrieval and chat agents, making Psy-Copilot easy to reproduce.

\subsubsection{Retrieval Argument Generation}
Based on two types of embedding index in Psy-COT, we adopt a two-stage retrieval argument generation for Psy-Copilot. As shown on the right side of Figure~\ref{fig:overview}, green arrows indicate dialogue retrieval pipeline for dialogue and blue arrows show retrieval progress for generated reasoning data. 

In the first step marked by the green arrow, the agent uses the conversation history to find similar dialogues in the Psy-COT dialogue index by comparing them using cosine similarity. Next, the agent reasons step-by-step based on the conversational history, to analyze the user's case and determine counseling strategies \cite{wei2022COT}. Based on the generated reasoning, the agent retrieves  relevant counseling strategies in the COT index.

We filter the nodes and their neighboring nodes obtained from the results of COT retrieval and dialog retrieval, prioritizing overlapping nodes. We concatenate the content of dialogue nodes into few-shot examples, and COT contents into instructions. Finally, the model generates psychological counseling responses based on retrieval information.

\subsubsection{Demo UI}

We have constructed an online website to provide an interactive experience with Psy-Copilot and showcase the Psy-COT graph. During each dialogue round, we dynamically retrieve and display related sub-graphs for the therapist. Sub-graphs make the logic chain behind generation easier to understand and trace. 

\subsubsection{Objective Evaluation}
Our agent can generate multi-turn dialogue. However, the interactive evaluation of multi-turn dialog is time-consuming and difficult to quantify. Inspired by Emobench~\cite{EmoBench2024}, we apply GLM4-9b~\cite{du2022glm} LLM for evaluating the emotional intelligence of the Psy-Copilot.

As shown in Table~\ref{tab:result}, we assessed the multi-turn dialogues with 4 metrics: Fluency (Flu.), Helpfulness (Hel.), Naturalness (Nat.), and Comforting Effectiveness (Com.). Detailed definitions of metrics and evaluation codes are available on our paper's website. Our experiments show that Psy-Agent outperforms baseline models across all four metrics,
\begin{table}[h]
\centering

    \begin{tabular}{lcccc}
    \hline
     \textbf{Model} & \textbf{Flu.} & \textbf{Hel.} & \textbf{Nat.} & \textbf{Com.} \\
    \hline
    Deepseek-V2 & 8.2 & 7.2 & 6.7 & 7.6 \\
    Psy-Copilot-Dialog & 8.5 & 7.4 & \textbf{7.2} & \textbf{8.2} \\
    Psy-Copilot-COT & \textbf{8.6} & \textbf{7.5 }& 7.0 & 7.9 \\
    \hline
    \end{tabular}
\caption{Human evaluation focuses on fluency, helpfulness, naturalness, and comforting effectiveness.}

\label{tab:result}
\end{table}
\section{Conclusion}
We demo Psy-COT and Psy-Copilot, a visual graph and a conversational agent aimed at providing traceable and reliable psychological counseling. Psy-Copilot can act as a helpful AI assistant for psychotherapists. We open-source the pipeline for graph construction and deploy an interactive counseling conversation platform. Our goal is to demonstrate the AI's response crafting process, cognitive patterns, and interaction techniques with patients, thereby fostering a sense of trust in the AI system among therapists. We aim to enhance the synergy between AI and therapists, creating a harmonious and balanced relationship among AI, therapists, and patients.

\bibliography{aaai25}
\end{document}